\documentclass{article}

\usepackage{microtype}
\usepackage{graphicx}
\usepackage{soul, color}
\usepackage{xcolor}
\usepackage{subfigure}
\usepackage{amsfonts}
\usepackage{amsmath}
\usepackage{mathtools}
\usepackage{graphics}
\usepackage{mathptmx}
\usepackage{enumitem}
\newtheorem{theorem}{Theorem}[section]

\newtheorem{lemma}[theorem]{Lemma}
\newtheorem{defi}[theorem]{Definition}
\newtheorem{assu}[theorem]{Assumption}
\newtheorem{propo}[theorem]{Proposition}
\usepackage{mwe}
\usepackage{multirow}
\usepackage[ruled,vlined]{algorithm2e}
\DeclareMathAlphabet{\altmathcal}{OMS}{cmsy}{m}{n}
\usepackage{booktabs} 
\usepackage{hyperref}

\usepackage[accepted]{icml2021}

\icmltitlerunning{Learning Diverse-Structured Networks for Adversarial Robustness}

\begin{document}

\twocolumn[
\icmltitle{Learning Diverse-Structured Networks for Adversarial Robustness}



\icmlsetsymbol{equal}{*}

\begin{icmlauthorlist}
\icmlauthor{Xuefeng Du}{equal,hkbu,wisc}
\icmlauthor{Jingfeng Zhang}{equal,riken}
\icmlauthor{Bo Han}{hkbu}
\icmlauthor{Tongliang Liu}{syd}
\icmlauthor{Yu Rong}{tencent}\\
\icmlauthor{Gang Niu}{riken}
\icmlauthor{Junzhou Huang}{tencent}
\icmlauthor{Masashi Sugiyama}{riken,tykyo}
\end{icmlauthorlist}

\icmlaffiliation{hkbu}{Hong Kong Baptist University}
\icmlaffiliation{wisc}{University of Wisconsin-Madison}
\icmlaffiliation{riken}{RIKEN}
\icmlaffiliation{syd}{University of Sydney}
\icmlaffiliation{tencent}{Tencent AI Lab}
\icmlaffiliation{tykyo}{University of Tokyo}

\icmlcorrespondingauthor{Bo Han}{bhanml@comp.hkbu.edu.hk}
\icmlcorrespondingauthor{Tongliang Liu}{tongliang.liu@sydney.edu.au}
\icmlcorrespondingauthor{Gang Niu}{gang.niu@riken.jp}

\icmlkeywords{Machine Learning, ICML}

\vskip 0.3in
]



\printAffiliationsAndNotice{\icmlEqualContribution} 

\begin{abstract}
In \emph{adversarial training}~(AT), the main focus has been the objective and optimizer while the model has been less studied, so that the models being used are still those classic ones in \emph{standard training}~(ST).
Classic \emph{network architectures}~(NAs) are generally worse than searched NAs in ST, which should be the same in AT. In this paper, we argue that NA and AT cannot be handled independently, since given a dataset, the optimal NA in ST would be \emph{no longer optimal} in AT.
That being said, AT is time-consuming itself; if we directly search NAs in AT over large \emph{search spaces}, the computation will be practically infeasible.
Thus, we propose a \emph{diverse-structured network}~(DS-Net), to significantly reduce the size of the search space: instead of low-level operations, we only consider predefined \emph{atomic blocks}, where an atomic block is a time-tested building block like the residual block.
There are only a few atomic blocks and thus we can weight all atomic blocks rather than find the best one in a searched block of DS-Net, which is an essential trade-off between \emph{exploring} diverse structures and \emph{exploiting} the best structures.
Empirical results demonstrate the advantages of DS-Net, i.e., weighting the atomic blocks.


\end{abstract}


\section{Introduction}
Safety-critical areas, such as autonomous driving, healthcare and finance, necessitate deep models to be adversarially robust and generalize well~\cite{GoodfellowSS14}. Recently, \textit{adversarial training (AT)} has been shown effective for improving the robustness of different models~\cite{MadryMSTV18}. Compared with \textit{standard training (ST)} on natural data, AT is a new training scheme, which generates adversarial examples on the fly and employs them to update model parameters~\cite{MadryMSTV18,ZhangYJXGJ19,abs-2002-11242,WongRK20,pang2021bag}.
\begin{table}[]
    \centering
     \tabcolsep 0.04in\renewcommand\arraystretch{0.745}{\small{}}%
    \footnotesize
    \caption{\small{Performance misalignment for different NA in terms of robustness (PGD-20) and standard accuracy. Models which perform better after ST are not necessarily more robust after AT. AdaRKNet denotes dynamic system-inspired network~\cite{KimCPKHK20}. AT is performed by PGD-10 with a perturbation bound of 0.031. Robustness is evaluated with the same perturbation bound of 0.031.}}
      \vspace{0.5em}
    \begin{tabular}{c|cc|cc}
    \hline
       Model  & Standard Acc. & Ranking & Robustness & Ranking
       \\
       \hline
       WRN-28-10& 0.9646&1&0.4872&3\\
       ResNet-62 &0.9596&2&  0.4855&4\\
       DenseNet-121  & 0.9504&3& 0.4993&2\\
          MobileNetV2  & 0.9443&4&0.4732&6 \\
          AdaRKNet-62& 0.9403&5&0.5016&1\\
          ResNet-50 &0.9362&6& 0.4807&5\\
         \hline
    \end{tabular}
    \label{tab:misalignment}
\end{table}

The research focus of AT has mainly been the \textit{objective} and \textit{optimizer} while the \textit{model} has been less studied. Therefore, it is urgent to explore the influence of network architectures (NAs) for adversarial robustness. Some emerging studies imply that the classic human-designed NAs (e.g., ResNet~\cite{HeZRS16}), specified for ST, may not be suitable for AT.
For example, \citet{lmjicml2020} and \citet{KimCPKHK20} argued that the forward propagation of ResNet can be explained as an explicit Euler discretization of an ordinary differential equation (ODE), which leads to unstable predictions given perturbed inputs. They proposed to change NAs according to more stable numerical schemes and thus obtained higher model robustness. Meanwhile, \citet{smmothadv} discovered that the ReLU activation function weakens AT due to its non-smooth nature. They replaced ReLU with its smooth approximations to improve robustness.

In addition, we show in Tab.~\ref{tab:misalignment} a \textit{misalignment phenomenon} for different NA in terms of their robustness after AT and standard accuracy after ST. This phenomenon further demonstrates a fact that \textit{manually-crafted NAs for ST may not be suitable for AT}. Specifically, among various architectures, a clear trend can be observed that models which perform better in terms of standard accuracy may not be more robust. Moreover, a newly-designed AdaRKNet-62~\cite{KimCPKHK20} has the \emph{biggest misalignment}, which inspires us to rethink NAs for AT. Namely, all intriguing results suggest improving adversarial robustness requires carefully modified structures. Nevertheless, designing an optimal architecture for AT is still a challenging problem.

One straightforward remedy is to search robust NAs \cite{GuoYX0L20,abs-2009-00902}, where the key to success is \textit{exploring} diverse structures. However, it suffers from computational inefficiency and is sometimes not effective. Specifically, AT is inherently time-consuming, and searching NAs over large spaces with AT drastically scales up the computation overhead~\cite{GuoYX0L20}. Besides, searching over a large space (denoted as the search phase) requires pruning less useful operations and retraining the model from scratch (denoted as the evaluation phase), which naturally leads to an optimization gap between these two phases. As claimed in \citet{optgap}, the search phase seeks to optimize a large network, but a well-optimized large network does not necessarily produce high-quality sub-architectures. Thus, the searched architecture may not always be more robust or generalizing better. This motivates us to find an effective architecture for AT which is easy to build and efficient to train while encouraging flexible NA exploration.

\begin{figure*}[t]
    \centering
    \includegraphics[width=1.0\textwidth]{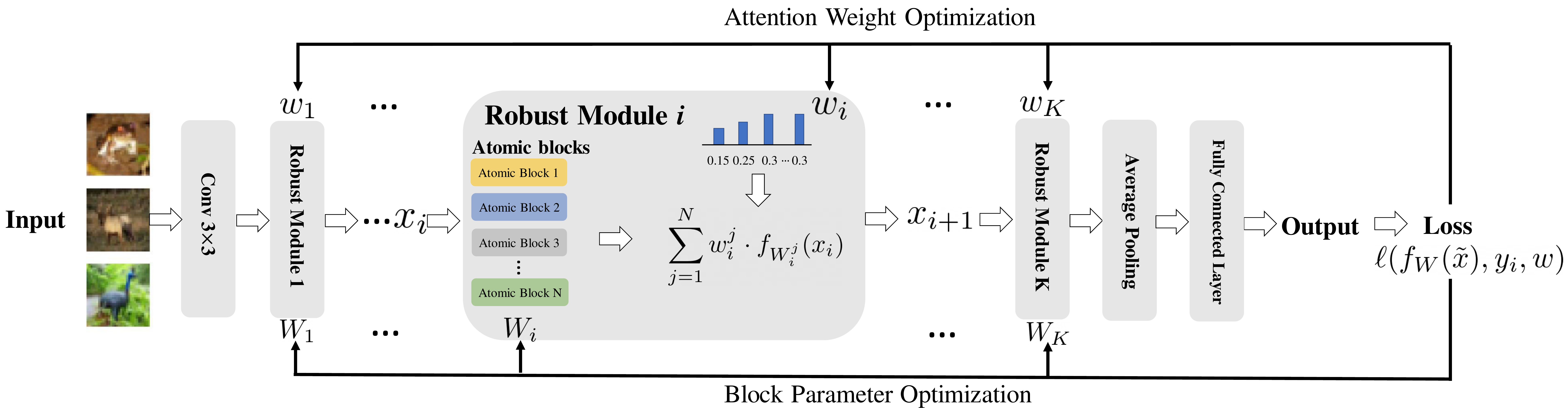}
    \caption{The network structure of DS-Net, where $\tilde{x}$ means the perturbed data and $f_W(\cdot)$ denotes the DS-Net with parameter $W$. The output from the $i$-th module $x_i$ goes through every atomic block and outputs $x_i^j$. The output $x_{i+1}$ is calculated by multiplying attention weights $w_i^j$ ($j=1,...,N$) and the corresponding outputs in a block-wise fashion. Here, $N$ is the number of atomic blocks, $K$ is the number of robust modules in sequence, and $W_i$ denotes the block parameter for the $i$-th module.}
    \label{fig:overview}
\end{figure*}

In this paper, we introduce a novel network design strategy which trades off \emph{exploring} diverse structures and \emph{exploiting} the best structures. Concretely, we propose a Diverse-Structured Network (DS-Net) as a novel solution to the trade-off (see Fig.~\ref{fig:overview}). Specifically, DS-Net consists of a sequence of modules. To significantly reduce the fine-grained search space, each module contains a few off-the-shelf time-tested building blocks (i.e., predefined atomic blocks in Section~\ref{sec:atomic-block}), which are either human-designed or search-based blocks that can be flexibly chosen. To encourage structure exploration, besides block parameters, we introduce a set of learnable attention weights to weight the outputs of these atomic blocks rather than finding the best one. The weights are concurrently optimized with the block parameters by the robust training objective and are fixed for evaluation.

Our end-to-end design strategy is analogous to manual design, which operates on a fixed set of atomic blocks but leverages attention weights to flexibly explore their relationship. It is different from searching robust NAs that determines the local structures inside each block by two-stage training. Additionally, the structure of DS-Net is consistent during AT, which does not require the operation of pruning that causes the optimization gap during searching NAs. Thus, our main contributions are summarized as follows:
\vspace{-3mm}
\begin{itemize}
\item We propose a novel DS-Net that trades off \emph{exploring} diverse structures and \emph{exploiting} the best structures. DS-Net remains the computational efficiency and effectiveness, which are limited in existing methods. 
\vspace{-3mm}
\item DS-Net allows for a flexible choice among powerful off-the-shelf atomic blocks including human-designed and search-based blocks, which are easy to understand, build, and robustify/generalize well.
\vspace{-0.5mm}
\item DS-Net learns attention weights of predefined atomic blocks based on the objective of AT. It empirically performs better than powerful defense architectures with less parameters on CIFAR-10 and SVHN.
\end{itemize}

\section{Related Work}
\textbf{Adversarial defense.}
Existing literature on the adversarial defense of neural networks can be roughly divided into two categories, namely certified robustness \cite{NeurIPS:Tsuzuku+etal:2018,ZhangL19} and empirical robustness~\cite{CaiLS18,MadryMSTV18,DBLP:conf/iclr/ZhangCXGSLBH20}. The former one focuses on either training provably robust models~\cite{WongK18} or obtaining certified models via random smoothing \cite{CohenRK19}, but often with a limited robustness compared to the latter approach. Empirical approaches usually rely on different techniques, such as input transformation \cite{DziugaiteGR16}, randomization \cite{XieWZRY18} and model ensemble \cite{LiuCZH18}.

However, most of them are evaded by adaptive attacks \cite{AthalyeC018}, while the most effective approaches till now are AT \cite{MadryMSTV18} and its variants \cite{ZhangYJXGJ19,0001ZY0MG20}. Based on it, many improvements were proposed, e.g., by metric learning \cite{LiYZZ19}, self-supervised learning \cite{NaseerKHKP20}, model-conditional training~\cite{WangCGH0W20}, weight perturbation~\cite{WuX020}, generative models \cite{WangY19} and semi-supervised learning \cite{abs-1906-00555}. Besides, several works attempted to speed up AT, such as computation reuse \cite{ZhangZLZ019}, adaptive inner maximization steps \cite{WangM0YZG19,abs-2002-11242} and one-step approximation \cite{WongRK20,SB20}. Note DS-Net improves AT from the viewpoint of network structure.

Several works attempted to improve adversarial robustness by diversity \cite{PangXDCZ19,abs-2002-05999,AbbasiRGB20,abs-1901-09981}, but neither of them focused on learning diverse-structured networks. 

\textbf{Robust network architecture.}
To obtain robust NAs, researchers developed smooth activation functions~\cite{smmothadv}, channel activation suppressing~\cite{anonymous2021improving}, dynamical system-inspired networks~\cite{lmjicml2020,KimCPKHK20}, model ensemble~\cite{WangSO19}, sparse coding~\cite{sparsecoding} and regularization~\cite{BuiLZMDAP20,RahnamaNR20} to enhance robustness. Besides, several works explored searching robust architectures~\cite{CubukZSL18,LiDWX20,hosseini2020dsrna,abs-1906-11667,abs-2011-09820,ChenZXGLJD20,abs-2012-11835}. Note DS-Net attempts to trade off \emph{exploring} diverse structures and \emph{exploiting} the best structures, which is orthogonal to these approaches.



\section{Proposed Approach}
\subsection{Preliminaries}
In this section, we briefly introduce the background for AT.

\textbf{Standard AT.} For each input $x$, let the input feature space $\altmathcal{X}$ with the infinity distance metric $d_{\mathrm{inf}}\left(x, x^{\prime}\right)=\left\|x-x^{\prime}\right\|_{\infty}$ be $(\altmathcal{X}, d_{\infty})$, the closed ball of radius $\epsilon>0$ centered at $x$ in $\altmathcal{X}$ be $\altmathcal{B}_{\epsilon}[x]=\left\{x^{\prime} \in \altmathcal{X} \mid d_{\inf }\left(x, x^{\prime}\right) \leq \epsilon\right\}$, and the function space be $\altmathcal{F}$. Given a dataset $S=\left\{\left(x_{i}, y_{i}\right)\right\}_{i=1}^{n}$ where $x_i\in\altmathcal{X}$ and $y_{i} \in \altmathcal{Y}=\{0,1, \ldots, C-1\}$, the objective function of the standard adversarial training \cite{MadryMSTV18} is
\vspace{-0.5em}
\begin{equation}
    \min _{f_W \in \altmathcal{F}} \frac{1}{n} \sum_{i=1}^{n}\left\{\max _{\tilde{x} \in \altmathcal{B}_{\epsilon}\left[x_{i}\right]} \ell\left(f_{W}(\tilde{x}), y_{i}\right)\right\},
    \label{eq:at_vanilla}
\end{equation}
where $\tilde{x}$ is the adversarial data centered at $x$ within the $\epsilon$-ball, $f_{W}(\cdot): \altmathcal{X} \rightarrow \mathbb{R}^{C}$ is a score function with parameters $W$ and $l(\cdot): \mathbb{R}^{C} \times \altmathcal{Y} \rightarrow \mathbb{R}$ is the loss function that is composed of a base loss $\ell_{\mathrm{B}}: \Delta^{C-1} \times \altmathcal{Y} \rightarrow \mathbb{R}$ (e.g., the cross-entropy loss) and an inverse link function $\ell_{\mathrm{L}}: \mathbb{R}^{C} \rightarrow \Delta^{C-1}$ (e.g., the soft-max activation). Here $\Delta^{C-1}$ is the corresponding probability simplex. In other words, $\ell(f(\cdot), y)=\ell_{\mathrm{B}}\left(\ell_{\mathrm{L}}(f(\cdot)), y\right)$. Denote $x^{(0)}$ as the starting point and $\alpha>0$ as the step size, standard AT generates the most adversarial data by Projected Gradient Descent (PGD) as follows:
\begin{equation}
\small
\label{eq:pgd}
  x^{(t+1)}=\Pi_{\altmathcal{B}\left[x^{(0)}\right]}\left(x^{(t)}+\alpha \operatorname{sign}\left(\nabla_{x^{(t)}} \ell\left(f_{W}\left(x^{(t)}\right), y\right)\right)\right), \forall t \geq 0,
\end{equation}
until a certain stopping criterion is satisfied to get the adversarial data $\tilde{x}$. $\Pi$ is the projection operator. It then minimizes the classification loss on $\tilde{x}$, which is agnostic to NAs.

\textbf{TRADES.} To trade off natural and robust errors, \citet{ZhangYJXGJ19} trained a model on both natural and adversarial data and changed the min-max formulation as follows:
 \vspace{-0.5em}
\begin{equation}
    \min _{f_W \in \altmathcal{F}} \frac{1}{n} \sum_{i=1}^{n}\left\{\ell\left(f_{W}\left(x_{i}\right), y_{i}\right)+\beta \ell_{\mathrm{K L}}\left(f_{W}\left(\tilde{x}_{i}\right), f_{W}\left(x_{i}\right)\right)\right\},
     \vspace{-0.5em}
\end{equation}
where $\ell_{\mathrm{K L}}$ is the Kullback-Leibler loss. $\beta$ is a regularization parameter that controls the trade-off between standard accuracy and robustness. When $\beta$ increases, standard accuracy will decease while robustness will increase, and vice visa. Meanwhile, the adversarial examples are generated by
 \vspace{-0.5em}
\begin{equation}
    \tilde{x}_{i}=\arg \max _{\tilde{x} \in \altmathcal{B}_{\epsilon}\left[x_{i}\right]} \ell_{K L}(f_{W}(\tilde{x}), f_{W}(x)).
     \vspace{-0.5em}
    \label{eq:trades}
\end{equation}

\textbf{Friendly adversarial training.} Friendly AT~\cite{abs-2002-11242} is a novel formulation of adversarial training that searches for least adversarial data
(i.e., friendly adversarial data) minimizing the
inner loss, among the adversarial data that are confidently misclassified. It is
easy to implement by just stopping the most adversarial data searching algorithms such as PGD
(projected gradient descent) early. The outer minimization still follows Eq.~\eqref{eq:at_vanilla}. However, instead of generating adversarial data via inner maximization, friendly AT generates $\tilde{x}_i$ as follows:
\begin{equation}
\begin{aligned} \tilde{x}_{i}= & \underset{\tilde{x} \in \altmathcal{B}_{\epsilon}\left[x_{i}\right]}{\arg \min }  \ell\left(f_W(\tilde{x}), y_{i}\right) \\ & \text { s.t. } \ell\left(f_W(\tilde{x}), y_{i}\right)-\min _{y \in \altmathcal{Y}} \ell(f(\tilde{x}), y) \geq \rho, \end{aligned}
\end{equation}
where there is a constraint on the margin of loss values $\rho$ (i.e., the misclassification confidence). This constraint firstly ensures $\tilde{x}$ is misclassified and secondly ensures for $\tilde{x}$ the wrong
prediction is better than the desired prediction $y_i$ by at least $\rho$ in terms of the loss value.

There are other AT styles, such as misclassification-aware AT~\cite{0001ZY0MG20} and Fast AT~\cite{WongRK20}.

\subsection{Diverse-Structured Network}
The overview of our DS-Net is demonstrated in Fig.~\ref{fig:overview}, which starts with a stem layer (e.g., a convolutional layer for images) for feature transformation. It then stacks $K$ sequential robust modules and ends with an average pooling layer and a fully connected layer. Each module has $N$ atomic blocks and two sets of variables for optimization, namely the attention weights $w$ and the block parameters $W$. We denote the attention weight for the $j$-th atomic block at the $i$-th module as $w_i^j$, which is randomly initialized before training. The feature transformation function of the $j$-th block at the $i$-th module is denoted as $f_{W_i^j}(\cdot)$ with the parameter $W_i^j$. During AT, DS-Net alternates between adversarial data generation (with the attention weights fixed) and classification loss minimization. For convenience, the following contents are described in the context of standard AT.

\textbf{Forward propagation.} Denote $x_{i}$ as the output feature of the $(i-1)^\mathrm{th}$ module and $g(\cdot)$ as the first stem layer, given an input $x$ or its adversarial counterpart $\tilde{x}$, the forward propagation of DS-Net is formulated as follows:
 \vspace{-0.5em}
\begin{equation}
    x_{i+1} = \sum_{j=1}^{N} w_i^j\cdot f_{W_i^j}(x_{i}),x_1=g(\tilde{x}),
    \vspace{-0.5em}
    \label{eq:weighted_sum}
\end{equation}
where the output of each module is calculated as the weighted sum of outputs of different atomic blocks. Each atomic block in a robust module has the same number of input and output channels.

\textbf{Backward propagation.} During the backward propagation to generate adversarial examples, DS-Net fixes attention weights $w$ and uses PGD to generate adversarial data as
\begin{equation}
\small
\label{eq:pgd_ds}
  x^{(t+1)}=\Pi_{\altmathcal{B}\left[x^{(0)}\right]}\left(x^{(t)}+\alpha \operatorname{sign}\left(\nabla_{x^{(t)}} \ell\left(f_{W}\left(x^{(t)}\right), y, w\right)\right)\right), \forall t \geq 0,
\end{equation}
which is similar to Eq.~\eqref{eq:pgd} but with a set of attention weights. 

During classification loss minimization, the attention weights $w$ and the atomic block parameters $W$ are optimized by Stochastic Gradient Descent (SGD) as:
\begin{equation}
\begin{split}
    w^{\prime}&= w-\alpha_{w}\nabla_{w}\ell\left(f_{W}(\tilde{x}), y_{i}, w\right),\\
    W^{\prime} &=W-\alpha_{W}\nabla_{W}\ell\left(f_{W}(\tilde{x}), y_{i}, w\right),\\
    \end{split}
    \label{eq:descent_ds}
\end{equation}
where $\alpha_{w},\alpha_{W}>0$ are the learning rate for the two set of variables.

Under standard AT, the minimax formulation is changed as
\begin{equation}
    \min _{w, f_W \in \altmathcal{F}} \frac{1}{n} \sum_{i=1}^{n}\left\{\max _{\tilde{x} \in \altmathcal{B}_{\epsilon}\left[x_{i}\right]} \ell\left(f_W(\tilde{x}), y_{i}, w\right)\right\},
    \label{eq:minimax_ds}
\end{equation}
where $f_W$ means DS-Net with the full set of atomic block parameters $W$. $W$ is simultaneously optimized with the block weights $w$ by the classification loss on the generated adversarial data. Therefore, DS-Net is able to automatically learn to weight different atomic blocks so as to improve architecture exploration and diversity.  The training and evaluation outline of DS-Net is presented in Algorithm \ref{algo:algo}.
 \vspace{-1em}
\begin{algorithm}[H]
\SetAlgoLined
\textbf{Input:} input data $x\in\altmathcal{X}$ with label $y\in\altmathcal{Y}$, model $f_W$ with block parameters $W$, loss function $\ell$, maximum PGD steps $K$, perturbation bound $\epsilon$, step size $\alpha$, and randomly initialized attention weights $w$.\\
\textbf{Output:} learned model $f_W$ and attention weights $w$.\\
\While{not eval}{
Step 1: Fix $w$ and $W$, generate $\tilde{x}$ by Eq.~\eqref{eq:pgd_ds}.\;
 
Step 2: Update $w$ and $W$ by Eq.~\eqref{eq:descent_ds}.\;
 }
  \While{eval}{
Step 3: Fix $w$ and $W$, generate $\tilde{x}$ by Eq.~\eqref{eq:pgd_ds}.\;

Step 4: Calculate output by Eq.~\eqref{eq:weighted_sum} and report accuracy.\;

 }
 \caption{Diverse-Structured Network.}
 \label{algo:algo}
\end{algorithm}
 \vspace{-1em}
\subsection{Predefined Atomic Blocks}\label{sec:atomic-block}
\begin{figure}[t]
    \centering
    \includegraphics[width=0.5\textwidth]{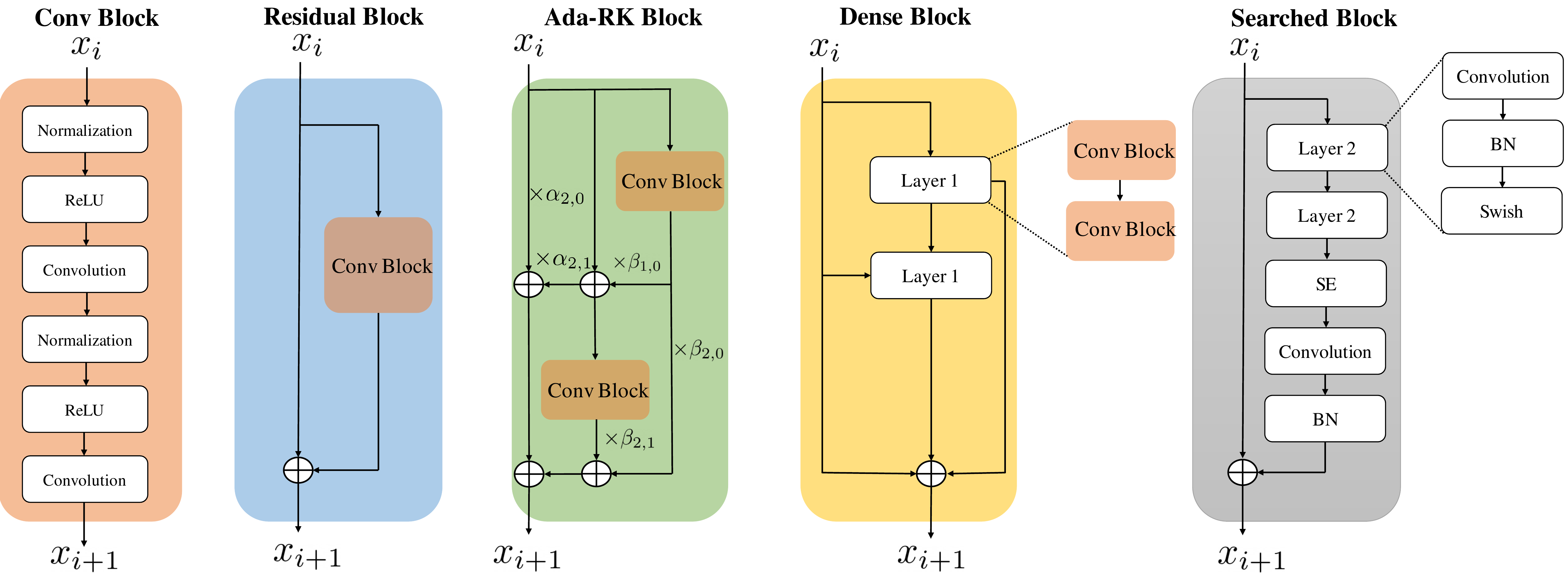}
      \vspace{-1em}
    \caption{The atomic blocks (except for the Conv block) in DS-Net. SE means the Squeeze-and-Excitation Layer \cite{HuSS18}. Swish is an activation function~\cite{RamachandranZL18}.}
    \label{fig:basis_block}
      \vspace{-1em}
\end{figure}

DS-Net allows for a flexible choice of the atomic blocks. We implement DS-Net by using four powerful atomic blocks in Fig.~\ref{fig:basis_block}, which are either human-designed, such as the residual block~\cite{HeZRS16}, dense block~\cite{HuangLMW17} and Adaptive Runge Kutta (Ada-RK) block~\cite{KimCPKHK20}, or searched-based block~\cite{TanCPVSHL19}. Most blocks have been theoretically or empirically validated to improve ST instead of AT. Although Ada-RK aims at AT, its generalization ability is not satisfactory~\cite{KimCPKHK20}. To trade off robustness and generalization, we simultaneously leverage four predefined atomic blocks via the learnable attention weights. Note that there are other potential candidates except the above four predefined blocks, which is a promising future work beyond the scope of our study.

Importantly, using atomic blocks avoids costly structure search for
AT while exploiting powerful architectures in literature. Besides, building a robust model by these blocks is easier while
retaining sufficient diversity, which is important for AT.

\subsection{Robustness Analysis}
We provide a robustness analysis of our DS-Net. Previous works~\cite{WengZCYSGHD18,HeinA17} usually connected Lipschitz smoothness w.r.t.~the input
with network robustness, which suggests that a small input perturbation will not lead to large change of the output. Formally, we give its definition.
\begin{defi}
\textup{\cite{Lipschitz} Given a model $f$ and a perturbation $\delta$ within the $\epsilon$-ball of the input $x$. The Lipschitz smoothness of $f$ is represented as}
\begin{equation}
    |f(x)-f(x+\delta)| \leq L_f\|\delta\|_{p}\leq L\epsilon,
\end{equation}
\textup{where $ p\ge0$ is the norm of interest and $L_f$ is the Lipschitz constant. Thus, a robust model holds a small value of $L_f$.}
\end{defi}

We fist provide Lemma~\ref{lemm:decomp} that decomposes the global Lipschitz constant into the value for each atomic block and then propose our main proposition (Proposition~\ref{the:robustness}).
\begin{lemma}
Denote the attention weight and the Lipschitz constant of the $j$-th block at the $i$-th module for DS-Net $f$ as $w_i^j$ and $L_i^j$, and the Lipschitz constant of the $j$-th block in the common network architecture $\hat{f}$ as $L_j$. If the number of layers and atomic blocks in DS-Net are $K$ and $N$, the Lipschitz constants can be decomposed as 
\begin{equation}
    L_f =\prod_{i=1}^K\sum_{j=1}^{N}w_i^j\cdot L_i^j, \quad L_{\hat{f}}=\prod_{j=1}^{NK}L_j.
\end{equation}
    \label{lemm:decomp}
\end{lemma}

\begin{propo}
Denote the Lipschitz constants for DS-Net with learnable or fixed attention weights as $L_f$ and $L^\prime_{f}$, and those of the common network architectures as $L_{\hat{f}}$, we get $L_f\leq L_{\hat{f}}$ and $L_f\leq L^\prime_{f}$. 
\label{the:robustness}
\end{propo}

\begin{figure}[t]
    \centering
    \includegraphics[width=0.47\textwidth]{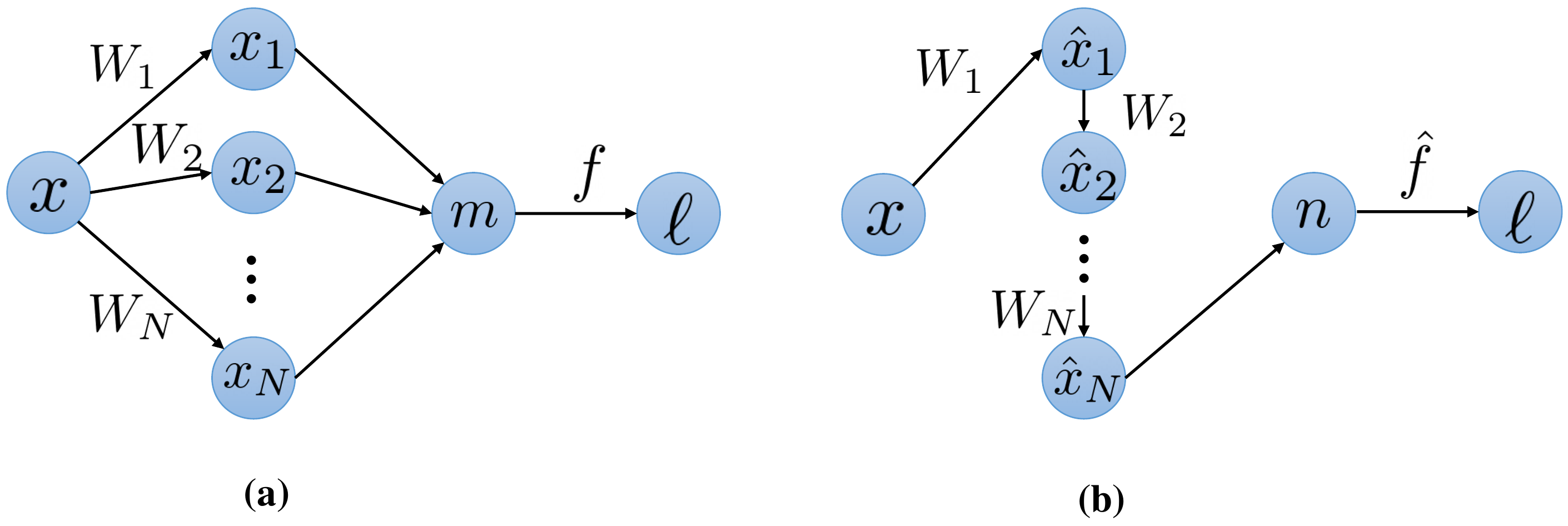}
    \vspace{-1em}
    \caption{Two different architectures. (a) The robust module for DS-Net. (b) The common network architecture, which covers many human-designed models and searched models. $x_i,\hat{x}_i$ denote the intermediate outputs (atomic block outputs for DS-Net). $f,\hat{f}$ denote the training objective functions. $m,n$ denote the outputs.}
    \label{fig:structure_analysis}
    \vspace{-1em}
\end{figure}

\textbf{Remark:}~From Proposition~\ref{the:robustness}, we conclude two results: 1) A common network architecture $\hat{f}(\cdot)$ (Fig.~\ref{fig:structure_analysis}(b)) with the same number of parameters as DS-Net (Fig.~\ref{fig:structure_analysis}(a)) always holds a larger Lipschitz constant $L_{\hat{f}}$. 2) The Lipschitz constant $L_f$ of DS-Net with learnable attention weights is smaller than that of DS-Net with an arbitrary fixed set of attention weights. The proof can be found in Appendix A. 

\subsection{Convergence Analysis}
We provide a convergence analysis of DS-Net for solving the min-max
optimization problem in Eq.~\eqref{eq:minimax_ds}. Due to the nonlinearities in DNNs such as ReLU~\cite{NairH10} and pooling operations, the exact assumptions of Danskin’s
theorem~\cite{danskin} do not hold. Nonetheless, since adversarial training only computes approximate maximizers of the inner problem, we can still provide a theoretical guarantee of convergence. The several necessary assumptions for the analysis are given as follows.

\begin{assu} Let $L_{WW},L_{Wx},L_{xW}$ be positive constants, the function $f_W(x,y)$ satisfies the gradient
Lipschitz conditions as follows
\begin{equation}
    \begin{array}{l}
\sup _{x}\left\|\nabla_{W} f_W( x,y)-\nabla_{W} f_{W^{\prime}}\left(x,y\right)\right\|_{2} \leq L_{WW}\left\|W-W^{\prime}\right\|_{2}; \\
\sup _{W}\left\|\nabla_{W} f_W( x,y)-\nabla_{W} f_{W}\left(x^{\prime},y\right)\right\|_{2} \leq L_{W x}\left\|x-x^{\prime}\right\|_{2}; \\
\sup _{x}\left\|\nabla_{x} f_W( x,y)-\nabla_{x} f_{W^{\prime}}\left(x,y\right)\right\|_{2} \leq L_{x W}\left\|W-W^{\prime}\right\|_{2}.
\end{array}
\end{equation}
\vspace{-1em}
\label{ass:1}
\end{assu}
Assumption~\ref{ass:1} requires that the loss function satisfies the Lipschitzian smoothness conditions. Despite the non-smoothness of the ReLU activation function, recent studies~\cite{Allen-ZhuLS19,DuLL0Z19,DBLP:journals/corr/abs-1902-01384} justify the loss function of overparamterized networks are semi-smooth. Thus this assumption is satisfied.
\begin{assu}
The variance of the stochastic gradient of $f_W(x)$ is bounded by a constant $\sigma^2>0$ as
\begin{equation}
\mathbb{E}\left[\left\|g\left(x_{k}, \xi_{k}\right)-\nabla f_W\left(x_{k}\right)\right\|^{2}\right] \leq \sigma^{2},
\end{equation}
where $g\left(x_{k}, \xi_{k}\right)$ is an unbiased estimator of $\nabla f_W\left(x_{k}\right)$ as $\mathbb{E}\left[g\left(x_{k}, \xi_{k}\right)\right]=\nabla f_W\left(x_{k}\right)$. $\xi_{k}, k \geq 1$ are random variables.
\label{ass:2}
\end{assu}
Assumption~\ref{ass:2} is commonly used in stochastic gradient based optimization algorithms~\cite{WangM0YZG19}.

Then we introduce the convergence analysis of non-convex optimization with the randomized stochastic gradient method~\cite{GhadimiL13a} as follows:
\begin{theorem}~\cite{GhadimiL13a}
Suppose the technical assumptions~\ref{ass:1} and~\ref{ass:2} hold. Let $f$ be $L$-smooth and non-convex function and $f^{*}$ be the optimal value of the optimization problem~\eqref{eq:minimax_ds}. Given repeated, independent accesses to stochastic gradients with variance bound $\sigma^2$, let SGD start with initial network parameters $W_0$, iterations $N>0$ and step size $\gamma_k<\frac{1}{L}$, then it converges to the following point by randomly choosing $W_k$ as the final output $W_R$ with probability $\frac{\gamma_k}{H}$. For $H=\sum_{k=1}^{N} \gamma_{k}$:
\begin{equation}
    \mathbb{E}\left[\left\|\nabla f_{W_{R}}\left(x\right)\right\|^{2}\right] \leq \frac{2\left(f_{W_{0}}\left(x\right)-f^{*}\right)}{H}+\frac{L \sigma^{2}}{H} \sum_{k=1}^{N} \gamma_{k}^{2},
    \vspace{-0.5em}
\end{equation}
where $f$ is the robust network in our case. 
\label{the:convergence_analysis}
\end{theorem}
\vspace{-0.5em}
From this theorem, we can see that the convergence speed and stability of the optimization problem heavily depend on the Lipschitz smoothness $L$ and the gradient variance $\sigma^2$ given the fixed number of iterations $N$ and SGD step size (i.e.,~learning rate) $\gamma_k$. 

In the following theorem, we explain that DS-Net has a smaller Lipschitz smoothness constant and gradient variance than common NAs. We also conduct an empirical analysis on these two factors to support our claim (Section~\ref{sec:eca}).

Due to the complexity of global Lipschitz smoothness, we instead use the block-wise
Lipschitz smoothness~\cite{BeckT13} in the following. Consider two architectures in Fig.~\ref{fig:structure_analysis} with the same number of parameters, then we have the following theorem.
\begin{theorem}
\label{the:convergence_bound}
Following Fig.~\ref{fig:structure_analysis}, let $\lambda_i$ be the largest eigenvalue of the network parameters $W_i,i=1,\ldots,N$. $f,\hat{f}$ are the objective functions for DS-Net and common network architectures, respectively. For any two possible assignments $W^1_i, W^2_i$ of $W_i$, the block-wise Lipschitz smoothness w.r.t. the network parameters and the gradient variance of the common network architectures are represented as
\begin{equation}
\small
\begin{split}
    \frac{\| \nabla_{W^1_i}\hat{f}- \nabla_{W^2_i}\hat{f}\|}{\|W^1_i-W^2_i \|} &\leq L_i\cdot\prod_{k=1}^{i-1} \lambda_{k},\\
    \mathbb{E}\| \nabla_{W_i}\hat{f}-\mathbb{E}\nabla_{W_i}\hat{f}\|^2&\leq N\sum_{k=i}^{N}\left(\frac{\sigma_{k}}{\lambda_{i}} \prod_{j=1}^{k} \lambda_{j}\right)^{2},
\end{split}
\end{equation}
by assuming the two properties of DS-Net satisfy: $\| \nabla_{W^1_i}f- \nabla_{W^2_i}f\| \leq L_{i}\left\|W^{1}_{i}-W^{2}_{i}\right\|$, $\mathbb{E}\left\|\nabla_{W_i}f-\mathbb{E}\nabla_{W_i}f\right\|^{2} \leq \sigma_{i}^{2}$. $N$ is the number of intermediate layers or atomic blocks.
\end{theorem}
\textbf{Remark:}~From Theorem~\ref{the:convergence_bound}, we conclude two results: 1) The common network architectures hold a Lipschitz smoothness constant, namely, $\prod_{k=1}^{i-1} \lambda_{k}$ times that of DS-Net. Note most of the largest eigenvalues of neural networks are bigger than~$1$ to prevent vanishing gradients~\cite{PascanuMB13}. Therefore, the convergence of Problem~\eqref{eq:minimax_ds} is slower than that of DS-Net. 2) The bound of the gradient variance in common network architectures is scaled up by the largest eigenvalue of network parameters and the network depth, which hurts the convergence speed and stability of AT. The proof can be found in Appendix B, which is adapted from~\citet{Shu0C20}. Overall, Theorem~\ref{the:convergence_bound} tells us DS-Net (Fig.~\ref{fig:structure_analysis} (a)) converges faster and more stably than common network architectures (Fig.~\ref{fig:structure_analysis} (b)).

\section{Experiments and Results}
In this section, we present empirical evidence to validate DS-Net on benchmarks with three AT styles. The code is available at~\url{https://github.com/d12306/dsnet}.
\subsection{Experimental Setting}
We evaluated DS-Net on CIFAR-10 and SVHN using: Projected Gradient Descent (PGD)~\cite{MadryMSTV18}, Fast Gradient Sign Method (FGSM)~\cite{GoodfellowSS14}, Carlini $\&$ Wagner (C$\&$W)~\cite{Carlini017} and AutoAttack (AA)~\cite{Croce020a}. We compared with human-designed models, such as ResNet~\cite{HeZRS16}, WideResNet~\cite{ZagoruykoK16}, IE-skips~\cite{lmjicml2020}, AdaRK-Net~\cite{KimCPKHK20} and SAT~\cite{smmothadv}. We also compared with searched NAs for AT, i.e., RobNet~\cite{GuoYX0L20}. We used three training styles, i.e., AT~\cite{MadryMSTV18}, TRADES~\cite{ZhangYJXGJ19} and MART~\cite{0001ZY0MG20}.

For CIFAR-10, during training, we set the perturbation bound $\epsilon$ to 0.031 and step size $\alpha$ to $0.007$ with 10 steps. We used SGD optimizer with a momentum of 0.9 and weight decay of 5e-4. The initial learning rate is 0.1. We trained for 120 epochs for standard AT and the learning rate is multiplied by 0.1 and 0.01 at epoch 60 and 90. For TRADES, we trained for 85 epochs and the learning rate is multiplied by $0.1$ at epoch 75. We tested the performance when the model is trained with regularization factor $\beta=1$ and $\beta=6$. For MART, we trained for 90 epochs and the learning rate is multiplied by $0.1$ at epoch 60. We set $\beta=6$. The batch size is set to 128. For SVHN, the step size is set to 0.003 with $\epsilon=0.031$. The training epochs including the epoch for learning rate decay is reduced by 20 for AT, TRADES and MART. We trained on one Tesla V100 and used mixed-precision acceleration by apex at $\textrm{O}1$ optimization level\footnote{\url{https://github.com/NVIDIA/apex}}. 
We select
all models 1 epoch after the 1st learning rate decay point following~\citet{DBLP:conf/icml/RiceWK20} because robust overfitting also happens for DS-Net. We have tried to use 1,000 images from
the training set as validation set to determine the
stopping point, which aligns with our selection point.

We used Adam optimizer~\cite{adam} with a learning rate of 1e-3 and a weight decay of 1e-3 to optimize the attention weights, which is then normalized by softmax function. The comparison with using other optimizers is shown in Appendix~E. We set the number of layers to 15 and the initial channel number to 20. We used two residual layers at the $\frac{1}{3}$ and $\frac{2}{3}$ of the total depth of the DS-Net to increase the channels by a factor of $k$ and 2, respectively. Meanwhile, the spatial size of the feature map is reduced by a half. We set $k=4/6$ and obtain a small and large DS-Net in our experiments, denoted as DS-Net-4/6-softmax.

The evaluation $\epsilon$ is set to $0.031$. We used PGD attack with 20 steps and CW attack with 30 steps. The step size is set to 0.031 for FGSM attack. For non-FGSM attack, we set the step size to 0.003 on TRADES while the evaluation step size on AT and MART is 0.008. We reported the best accuracy for comparison. Due to the complexity, we reported the accuracy of AA by randomly sampling $12.5\%$ of the test set. Each experiment is repeated by $3$ times with three random seeds. The results are averaged for comparisons.

\begin{table*}[ht]
    \centering
    \small
    \tabcolsep 0.04in\renewcommand\arraystretch{0.745}{\small{}}%
    \caption{Test accuracy on CIFAR-10 and SVHN using AT \cite{MadryMSTV18}. $\dag$ means the results by our implementation under the same setting. AA denotes the results of AutoAttack~\cite{Croce020a}. The perturbation bound $\epsilon$ is 0.031. DS-Net-4-softmax means the scale factor is 4 and block weights are normalized by the softmax activation. Improv.(\%) is calculated by comparing with the best baseline.}
    \begin{tabular}{c|c|ccccc}
    \hline
    \multicolumn{7}{c}{CIFAR-10}  \\
    \hline
         Defense Architecture&  Param (M)& Natural & FGSM&PGD-20&CW & AA\\
       \hline
         ResNet-50 \cite{HeZRS16}$\dag$ & 23.52 & 83.83$\pm$0.190 & 54.76$\pm$0.229& 48.07$\pm$0.222 &47.77$\pm$0.365 & 44.98$\pm$0.237\\
          WRN-34-10 \cite{ZagoruykoK16}$\dag$& 46.16&86.32$\pm$0.317 & 64.84$\pm$0.118&51.95$\pm$0.291& 50.65$\pm$0.339&   50.01$\pm$0.284 \\
           SAT-ResNet-50~\cite{smmothadv}$\dag$&23.52&73.59$\pm$0.164&57.50$\pm$0.287&48.44$\pm$0.105&46.56$\pm$0.291&44.11$\pm$0.370 \\
    SAT-WRN-34-10~\cite{smmothadv}$\dag$&46.16& 78.03$\pm$0.298&60.73$\pm$0.305&49.54$\pm$0.311&49.43$\pm$0.042&46.27$\pm$0.163\\
       IE-ResNet-50 \cite{lmjicml2020}$\dag$ & 22.41 & 84.49$\pm$0.111&55.00$\pm$0.229&  48.31$\pm$0.321& 48.04$\pm$0.392&43.27$\pm$0.138\\
         IE-WRN-34-10 \cite{lmjicml2020}$\dag$ & 48.24&84.23$\pm$0.200 & 63.28$\pm$0.222&52.61$\pm$0.316&49.36$\pm$0.501&51.24$\pm$0.251 \\
         AdaRK-Net~\cite{KimCPKHK20}$\dag$ &23.61&80.42$\pm$0.124&57.23$\pm$0.218&51.37$\pm$0.411&49.27$\pm$0.228&45.11$\pm$0.260 \\
         RobNet-large-v2 \cite{GuoYX0L20}$\dag$ &33.42&84.39$\pm$0.129& 59.21$\pm$0.311& 52.54$\pm$0.371&51.28$\pm$0.212&49.22$\pm$0.138  \\
         \hline
           DS-Net-4-softmax (ours)&20.78 &85.39$\pm$0.216&66.71$\pm$0.186&\textbf{54.14}$\pm$0.100&52.18$\pm$0.137& 49.98$\pm$0.199\\ 
           DS-Net-6-softmax (ours)& 46.35&\textbf{86.76}$\pm$0.125 &\textbf{67.03}$\pm$0.372 &53.59$\pm$0.211  &\textbf{53.28}$\pm$0.174 &\textbf{51.48}$\pm$0.191 \\
           \hline
           Improv.(\%)&-&0.51\%&3.38\%&2.91\%&3.90\%&0.47\%\\
          \hline
           \multicolumn{7}{c}{SVHN}  \\
           \hline
           ResNet-50~\cite{HeZRS16}$\dag$&23.52&90.02$\pm$0.213& 69.03$\pm$0.233& 47.23$\pm$0.177& 49.69$\pm$0.186& 44.11$\pm$0.029\\
           WRN-34-10~\cite{ZagoruykoK16}$\dag$&46.16&94.26$\pm$0.175& 75.15$\pm$0.310& 48.57$\pm$0.163& 50.08$\pm$0.271& 45.38$\pm$0.124\\
           \hline
            DS-Net-4-softmax (ours)&20.78 &95.53$\pm$0.172&\textbf{78.50}$\pm$0.278&49.53$\pm$0.301&48.73$\pm$0.101&46.21$\pm$0.222 \\
             DS-Net-6-softmax (ours)&46.35 &\textbf{95.96}$\pm$0.211&75.80$\pm$0.170&\textbf{50.89}$\pm$0.235&\textbf{50.12}$\pm$0.304&\textbf{48.09}$\pm$0.258 \\
           \hline
           Improv.(\%)&-&1.80\%&4.46\%&4.78\%&0.08\%&5.97\%\\
           \hline
    \end{tabular}
    \label{tab:at} 
\end{table*}

\subsection{Results on CIFAR-10 and SVHN}
We presented the results of AT and MART in Tabs.~\ref{tab:at} and~\ref{tab:fat}. Results of TRADES are in Appendix C. We made several observations. First, the robustness of DS-Net is consistently better than baselines, which achieves promising results with a much smaller amount of parameters, e.g., 54.14$\%$ under PGD-20 attack with only 20.78M parameters compared to 51.95\% for WRN-34-10 (46.16M) and 52.54\% for RobNet (33.42M) using AT. Second, if we increase the amount of parameters to the same level of WRN-34-10 by setting the factor $k=6$, DS-Net further improves its robustness ($53.28\%$ under CW attack). Meanwhile, DS-Net generalizes well (with a standard accuracy of 86.76\%) and also performs well in terms of ensembles of white-box and black-box attacks (see AutoAttack). Third, DS-Net shows its effectiveness across different datasets and training styles, which provides better robustness and generalization ability.
\begin{table*}[]
    \centering
    \small
    \tabcolsep 0.04in\renewcommand\arraystretch{0.745}{\small{}}%
     \caption{Evaluations (test accuracy) of deep models on CIFAR-10 and SVHN dataset using MART \cite{0001ZY0MG20}. $\dag$ means the results by our implementation. The perturbation bound $\epsilon$ is set to 0.031 for each architecture. The regularization factor $\beta$ is set to 6.}
    \begin{tabular}{c|c|ccccc}
    \hline
      \multicolumn{7}{c}{CIFAR-10}  \\
            \hline
      Defense Architecture&  Param (M)& Natural & FGSM&PGD-20&CW&AA\\
      \hline
      RobNet-large-v2 \cite{GuoYX0L20}$\dag$ &33.42&80.23$\pm$0.129&60.23$\pm$0.203&51.07$\pm$0.290& 48.37$\pm$0.365&48.14$\pm$0.317 \\
        WRN-34-10 \cite{ZagoruykoK16}$\dag$ & 46.16&78.59$\pm$0.221&62.50$\pm$0.355& 52.26$\pm$0.409 &49.75$\pm$0.517 & 49.96$\pm$0.531 \\
         IE-WRN-34-10 \cite{lmjicml2020}$\dag$ & 48.24&81.33$\pm$0.127&62.29$\pm$0.116&51.99$\pm$0.244&49.40$\pm$0.142& 50.07$\pm$0.246 \\
         \hline
          DS-Net-4-softmax(ours) & 20.76&79.51$\pm$0.137&63.03$\pm$0.241&54.29$\pm$0.376&50.25$\pm$0.229&49.79$\pm$0.256   \\
             DS-Net-6-softmax(ours) & 46.35&\textbf{81.64}$\pm$0.229&\textbf{66.40}$\pm$0.173&\textbf{55.23}$\pm$0.168&\textbf{51.48}$\pm$0.291&\textbf{52.74}$\pm$0.096   \\
             \hline
           Improv.(\%)&-&0.38\%& 6.09\%&5.68\%&3.48\%&5.44\%\\
          \hline
            \multicolumn{7}{c}{SVHN}  \\
            \hline
              WRN-34-10~\cite{ZagoruykoK16}$\dag$ & 46.16& 92.15$\pm$0.279&74.57$\pm$0.160&52.96$\pm$0.384&47.03$\pm$0.100&49.88$\pm$0.103\\
                \hline
              DS-Net-4-softmax (ours)&20.78 &92.39$\pm$0.172&73.88$\pm$0.263&\textbf{56.08}$\pm$0.326&48.00$\pm$0.298&\textbf{51.39}$\pm$0.206 \\
             DS-Net-6-softmax (ours)&46.35 & \textbf{93.77}$\pm$0.272&\textbf{76.23}$\pm$0.165&55.00$\pm$0.126&\textbf{48.84}$\pm$0.179&50.43$\pm$0.312\\
             \hline
           Improv.(\%)&-&1.76\%&2.23\%&5.89\%&3.85\%&3.03\% \\
             \hline
    \end{tabular}
    \label{tab:fat}
    \vspace{-1em}
\end{table*}

\subsection{Comparison with Block Ensemble}
\begin{figure}[t]
    \centering
    \vspace{-1em}
    \includegraphics[width=0.48\textwidth]{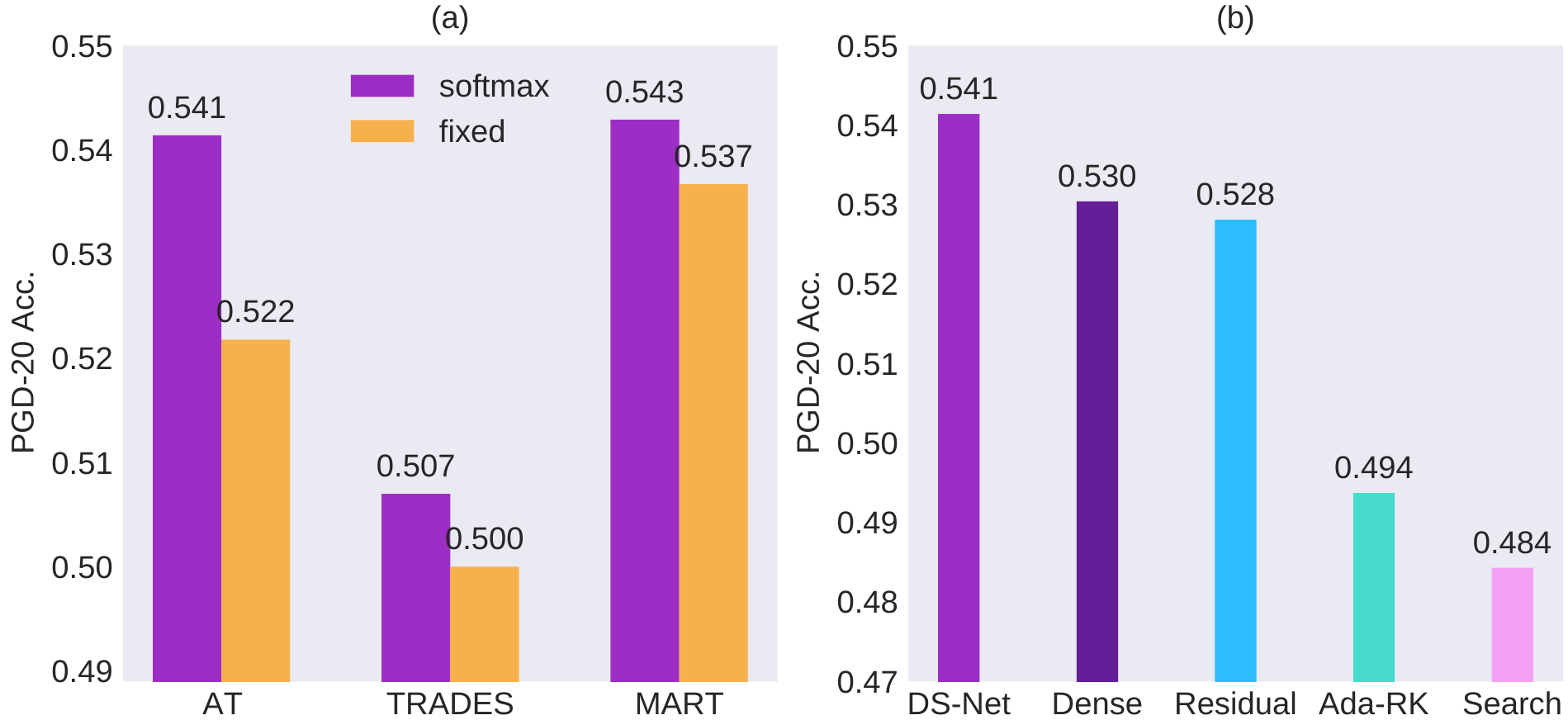}
    \caption{The results of block ensemble variants of DS-Net (under PGD-20 attack). (a) DS-Net with fixed and uniform attention weights. $\beta=1$ is used for TRADES. (b) DS-Net with the same atomic blocks and uniform ensemble. Standard AT is used.}
    \label{fig:comparison_with_ensemble}
    \vspace{-2em}
\end{figure}
\vspace{-0.8em}
We compared the robustness of DS-Net with two variants: 1) The attention weights are uniformly distributed among different atomic blocks and fixed during training. 2) The four atomic blocks are the same whose outputs are uniformly ensembled. We compared them with DS-Net on CIFAR-10 with factor $k=4$, which is shown in Fig.~\ref{fig:comparison_with_ensemble}.

The above figure highlights two factors---a) Learnability of attention weights and (b) The diversity of atomic blocks---matter for network robustness in DS-Net. Therefore, the improvement of DS-Net does not solely
come from a simple ensemble. Meanwhile, we found that the highest standard accuracy of these variants are lower than DS-Net, such as 85.31\% vs. 87.89\% for fixed attention weights and learnable weights in DS-Net on TRADES, which justifies that these two factors are also important for generalization.

\begin{figure}[t]
    \centering
    \includegraphics[width=0.47\textwidth]{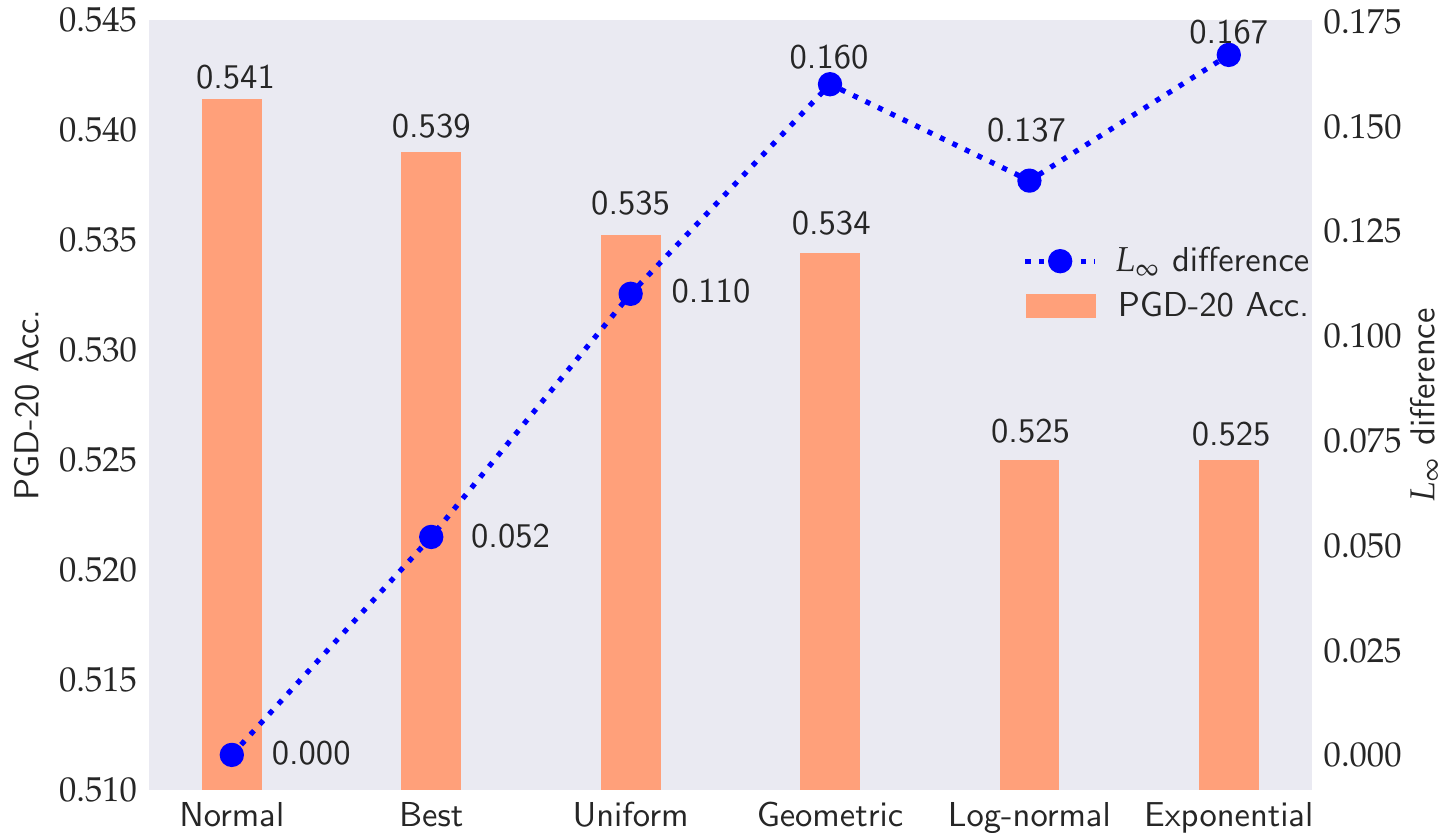}
    \caption{Comparative robustness of our DS-Net with different initializations (evaluated by PGD-20 attack). }
    \label{fig:comparison_initialization}
    \vspace{-1.5em}
\end{figure}
\subsection{Sensitivity to Weight Initialization}
We investigated the sensitivity of DS-Net to the initialization of the attention weights. Note the results in Tabs.~\ref{tab:at} and \ref{tab:fat} are reported using the normal distribution ($\mu=0, \sigma^2$=1), we further tested the uniform distribution (within $[0,1]$), log-normal distribution ($\mu=0, \sigma^2=1$), exponential distribution $\lambda=1$, geometric distribution ($p=0.5$). We also used the optimized attention weights after training for initialization. We compared these initializations using standard AT on CIFAR-10 with factor $k=4$, which are shown in Fig.~\ref{fig:comparison_initialization}.

From the above figure, DS-Net is not sensitive to different initializations, where the largest accuracy drop is $\leq2\%$. The optimized weight for initialization obtains the closet performance to the original one, which illustrates its superiority. Besides, we compared the $l_{\infty}$ difference between the learned weights initialized with the normal distribution and the others, which shows a similar trend to the PGD-20 Acc.

\begin{figure}[t]
    \centering
    \includegraphics[width=0.47\textwidth]{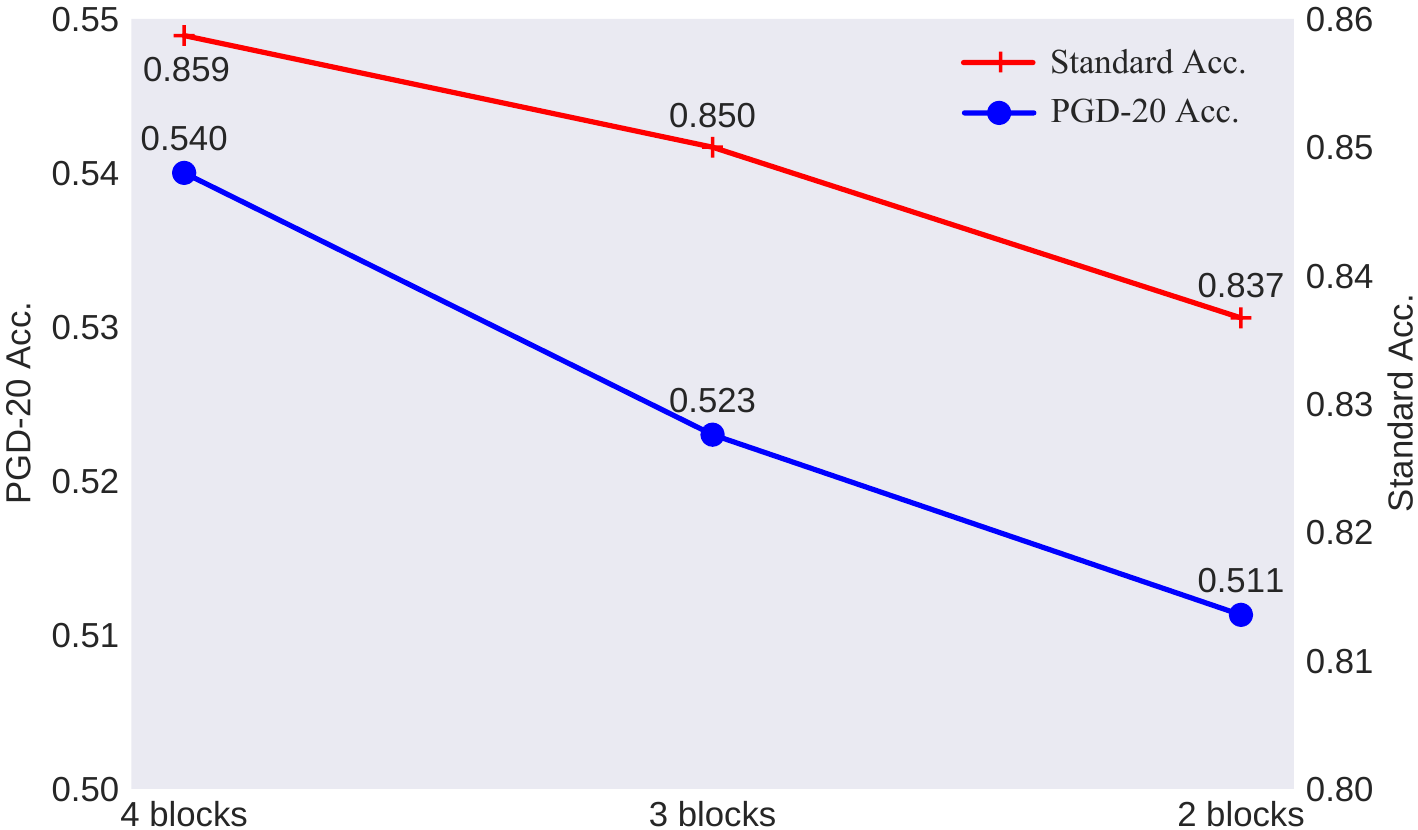}
    \vspace{-1em}
    \caption{The performance of DS-Net with atomic block space reduction. Results are reported using seed 0.}
    \label{fig:space_reduction}
    \vspace{-1em}
\end{figure}
\begin{figure*}[t]
    \centering
    \includegraphics[width=1.0\textwidth]{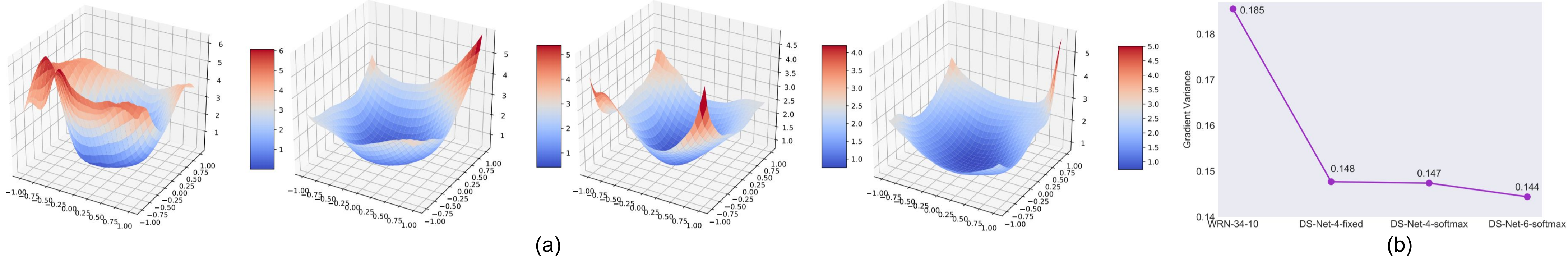}
    \caption{Empirical convergence analysis under TRADES ($\beta=6$). (a) 3D loss landscape of four models, i.e., WRN-34-10, DS-Net-4-fixed, DS-Net-4-softmax and DS-Net-6-softmax from left to right where ``fixed" means we use 0.25 as the weight for each atomic block. The height of the surface indicates the loss value. (b) Gradient variance for the same set of models.}
    \label{fig:landscape_variance}
     \vspace{-0.5em}
\end{figure*}
\vspace{-0.5em}

\subsection{Effect of Reducing Atomic Block Space}
To observe whether the number of atomic blocks matters to model performance, we trained our DS-Net under 2 and 3 atomic blocks for each robust module. Note that sampling 2 and 3 atomic blocks from 4 blocks has 6 and 4 options and we reported the average results. To ensure a fair comparison, we kept the same level of network parameters by changing the initial channel number across different models. We compared them with DS-Net-4-softmax using standard AT on CIFAR-10 with factor $k=4$, which are shown in Fig.~\ref{fig:space_reduction}.

From Fig.~\ref{fig:space_reduction}, the performance of DS-Net slightly decreases in terms of robustness and generalization ability with smaller atomic block space, which shows a carefully-designed block space with higher diversity is beneficial. A principled approach to block selection is a promising future work.

\subsection{Learned Attention Weight Visualization}
\begin{figure}[t]
    \centering
    \includegraphics[width=0.47\textwidth]{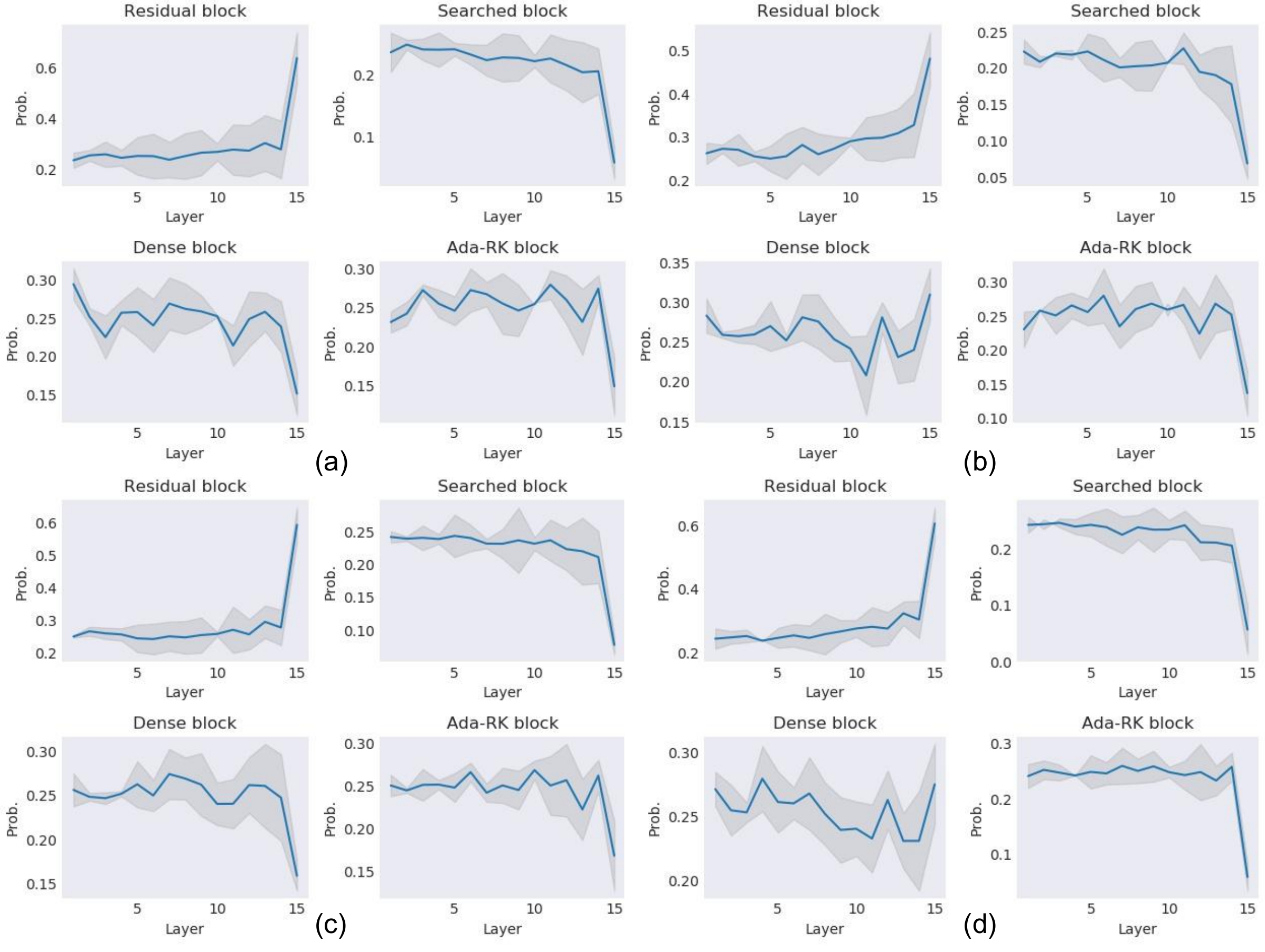}
    \caption{Visualization of the learned attention weights with variance in shaded area. (a) DS-Net-4-softmax under AT. (b) DS-Net-6-softmax under AT. (c) DS-Net-6-softmax under TRADES ($\beta=1$). (d) DS-Net-6-softmax under TRADES ($\beta=6$).}
    \label{fig:visualization}
    \vspace{-2.1em}
\end{figure}

To gain some insights on the location sensitivity of different atomic blocks, we visualized the learned attention weights from different layers in Fig.~\ref{fig:visualization}. We made several observations. First, the weight of atomic blocks is balanced in the former layers of DS-Net without obvious dominance, which implies that AT prefers a diverse structure. Second, the weight of residual block increases in the last layers, which shows DS-Net tends to tends to use the
cleaner feature representations by favoring residual blocks
in the later layers for classification. This phenomenon happens because the error of features
learned by the early layers accumulates less than that learned
by the later layers. Third, densely connected modules are more favorable in robust models. For instance, DS-Net-6-softmax pays more attention to the dense block compared to DS-Net-4-softmax, and DS-Net-6-softmax under TRADES ($\beta=6$) gives more weights to dense block than DS-Net-6-softmax under TRADES ($\beta=1$). Such findings align well with~\citet{GuoYX0L20}. The trends may guide us to design different blocks for different layers of a robust model.

\vspace{-0.5em}
\subsection{Empirical Convergence Analysis}
\label{sec:eca}
We showed in Theorem~\ref{the:convergence_analysis} that Lipschitz smoothness and gradient variance are important for convergence. In this section, we empirically verified these two properties of DS-Net. Due to the substantial budget for calculating the Hessian matrix of the objective function in order to measure the global Lipschitz smoothness~\cite{Nesterov04}, we followed \cite{Li0TSG18} and used the loss landscape to measure the local smoothness of models, which is visualized as $f(\alpha, \beta)=L\left(W^{*}+\alpha \delta+\beta \eta\right)$ ($\delta,\eta$ are two random directions and $W^{*}$ is the center point in the network parameter space). We compared the loss landscape among four models in Fig.~\ref{fig:landscape_variance}~(a), and found that DS-Net empirically smooths the loss landscape around the optimized parameters, which makes the convergence faster and more stable.

We computed the gradient variance of DS-Net by its definition $\mathbb{E}\| \nabla_{W_i}f-\mathbb{E}\nabla_{W_i}f\|^2$. To reduce computational cost, following~\citet{Shu0C20}, the gradients over batches are regarded as the full gradients and the expected gradient is calculated by averaging over batch gradients. The variance is calculated w.r.t. the last fully connected layer. The comparison among the same set of models is shown in Fig.~\ref{fig:landscape_variance}(b), where DS-Net holds a smaller gradient variance. Learnable attention weights and more network parameters are beneficial to convergence speed and stability.
\vspace{-0.5em}

\subsection{Results after Block Pruning}
To observe whether block pruning in search-based methods~\cite{abs-2009-00902} benefits DS-Net, we tested the performance of DS-Net by selecting 1,2,3 atomic blocks with higher probabilities after training, re-normalizing their weights and retraining the model from scratch by AT (The weights are fixed). We also tested 4 blocks with their optimized attention weights. To ensure a fair comparison, we kept the same level of network parameters by changing the initial channel number across different models. We conducted experiments using standard AT on CIFAR-10 with factor $k=4$, which are shown in Tab.~\ref{tab:pruning}.
\begin{table}[!h]
    \centering
    \small
     \tabcolsep 0.04in\renewcommand\arraystretch{0.745}{\small{}}%
         \caption{Model robustness and generalization ability with different number of pruned blocks. The standard deviations (Std.) are shown below the mean row (Acc.).}
    \begin{tabular}{c|ccccc}
    \hline
        Selected Blocks & 1&2&3&4&Ours \\
         \hline
      PGD-20 Acc. (\%)&42.03& 51.88 & 50.70& 52.19&\textbf{54.14}\\
     Std. & 0.339&
0.214& 0.213& 0.291& 0.100\\
      Standard Acc. (\%)& 78.05& 84.22& 84.30& 83.75& \textbf{85.39}\\
      Std.& 0.172& 0.238& 0.176& 0.118& 0.216 \\
         \hline
    \end{tabular}
    \vspace{-1em}
    \label{tab:pruning}
\end{table}

Tab.~\ref{tab:pruning} shows that block pruning is not suitable for DS-Net, since attention weights and block parameters are co-adapted together. Therefore, discarding parameters that cooperates well with attention weights will lead to the accuracy drop. Additionally, such block pruning reduces the structure diversity during retraining, which may also hurt the model robustness and generalization ability.

\vspace{-0.5em}
\section{Conclusion}
\vspace{-0.5em}
Orthogonal to studies from the view of objective and optimizer for AT, we focus on NAs and propose a Diverse-Structured Network (DS-Net) to improve model robustness. DS-Net trades off exploring diverse structures and exploiting the best structures. Specifically, it learns a set of attention weights over predefined atomic blocks, where attention weights are jointly optimized with network parameters by the robust training objective that encourages structure exploration. We theoretically demonstrate the advantage of DS-Net in terms of robustness and convergence, and empirically justify our DS-Net on benchmark datasets. In the future, we will improve DS-Net by studying different combination of atomic blocks to further improve model robustness.

\vspace{-1em}
\section*{Acknowledgements}
XFD and BH were supported by HKBU Tier-1 Start-up Grant and HKBU CSD Start-up Grant. BH was also supported by the RGC Early Career Scheme No. 22200720, NSFC Young Scientists Fund No. 62006202 and HKBU CSD Departmental Incentive Grant. TLL was supported by Australian Research Council Project DE-190101473. JZ, GN and MS were supported by JST AIP Acceleration Research Grant Number JPMJCR20U3, Japan. MS was also supported by the Institute for AI and Beyond, UTokyo.
\bibliography{arxiv}
\bibliographystyle{icml2021}



\appendix
\onecolumn

\renewcommand{\thesection}{A}
\section{Proof of Proposition 3.3}
Before we give the proof for Proposition 3.3, we first illustrate how we obtain Lemma 3.2.

\textit{Proof of Lemma 3.2} For the first module $f_1$ in DS-Net, the Lipschitz smoothness is represented as
\begin{equation}
    |f_1(x)-f_1(x+\delta)|\leq\sum_{j=1}^N w_{1}^j \cdot L_1^j \cdot  \|\delta\|_p,
\end{equation}
which is obtained by the definition of Lipschitz smoothness for each atomic block.
By composing different layers behind together, we get 
\begin{equation}
     |f_K(x)-f_K(x+\delta)|\leq \prod_{i=1}^{K}\sum_{j=1}^N w_{i}^j \cdot L_i^j \cdot  \|\delta\|_p,
\end{equation}
Therefore, the Lipschitz constant of DS-Net is decomposed as $L_f=\prod_{i=1}^{K}\sum_{j=1}^N w_{i}^j \cdot L_i^j$. Following the same decomposition process, the Lipschitz constant of the common network architecture is decomposed as $L_{\hat{f}}=\prod_{j=1}^{NK}L_j$ with the same number of network parameters.

\textit{Proof of Proposition 3.3} For the Lipschitz constant of the parameterized convolutional layers, we focus on the $L_2$ bounded perturbations. According to the definition of spectral norm, the Lipschitz constant of these layers is the spectral norm of its weight matrix. Mathematically, we get $L_i^j= \prod_{k=1}^{M} \|W_k \|_2$ where $M$ is the number of convolutional layers in the current block. The spectral norm is also the maximum singular value of $W_k$. According to~\citet{PascanuMB13}, we need $\|W_k\|_2\geq1$ in order to prevent vanishing gradient problem during adversarial training. Therefore, we get $L_i^j\geq1$ for all the blocks.

Comparing $L_f$ and $L_{\hat{f}}$ then degenerates to comparing $\sum_{j=1}^N w_{i}^j \cdot L_i^j$ for DS-Net and $\prod_{j=1}^{N}L_j$ for the common network architecture since we can reorder the blocks and do not change the Lipschitz constant. And we have the following comparison results
\begin{equation}
    \sum_{j=1}^N w_{i}^j \cdot L_i^j \leq max_{j} L_i^j \leq \prod_{j=1}^{N}L_j,
\end{equation}
which obtained by using the fact that $L_i^j\geq1$. Note that although the perturbation is $L_2$ bounded, the robustness against $L_{\infty}$ can be also achieved, as stated by~\cite{QianW19}.

Next, we prove DS-Net with a learnable attention weight is more robust than DS-Net with an arbitrary fixed set of attention weight. According to~\citet{CisseBGDU17,DBLP:journals/corr/abs-2004-01832}, the robust training objective $\ell(f(x+\delta,y,w))$ can be approximated by 
\begin{equation}
    \ell(f(x+\delta,y,w)) \approx  \ell(f(x,y,w)) + L\epsilon,
\end{equation}
which is the standard classification loss on the natural images plus a term that is linearly correlated with the global Lipschitz constant. 

Given a fixed number of network parameters, the standard classification loss function for the DS-Net with a learnable or fixed set of attention weights does not vary much. Therefore, adversarial training minimizes the global Lipschitz constant $L$ of DS-Net implicitly. Recall that the global Lipschitz constant $L$ is a function of the attention weight $w_i^j$ and the Lipschitz constant $L_i^j$ for each block, if we assume $w_i^j$ is learnable and adversarial training leads to a global minimization of $L$, then changing the optimized $w_i^j$ in DS-Net will cause the global Lipschitz constant $L$ to increase, which validates our claim $L_f\leq L^\prime_f$.

\renewcommand{\thesection}{B}
\section{Proof of Theorem 3.7}
The proof of Theorem 3.7 is inspired by~\citet{Shu0C20}.

\textit{Proof of Theorem 3.7} Following Fig.~3 in the main paper, we first explain how we obtain the bound of the block-wise Lipschitz constant for the common network architecture given a bounded block-wise Lipschitz constant of DS-Net. 

To begin with, the derivative w.r.t. the parameter $W_i$ in DS-Net is calculated as:
\begin{equation}
\nabla_{W_i}f=\nabla_{x_i}f x^T.
\label{eq:derivation}
\end{equation}
For the common network architecture, we get $\hat{x}_{i}=\prod_{k=1}^{i} W_{k} \boldsymbol{x}$. Similarly, the gradient w.r.t. the parameter matrix $W_i$ is calculated by the chain rule as
\begin{equation}
\begin{split}
        \nabla_{W_i}\hat{f} &= \sum_{k=i}^{N}\left(\prod_{j=i+1}^{k} W_j\right)^{T} \nabla_{\hat{x}_k}\hat{f}\left(\prod_{j=1}^{i-1} W_j x\right)^{T}\\
        &= \sum_{k=i}^{N}\left(\prod_{j=i+1}^{k} W_j\right)^{T} \nabla_{\hat{x}_k}\hat{f}x^T\left(\prod_{j=1}^{i-1} W_j \right)^{T}.\\
\end{split}
\end{equation}
Using the fact that $\nabla_{\hat{x}_k}\hat{f}=\nabla_{x_k}f$, we replace $\nabla_{\hat{x}_k}\hat{f}x^T$ with $\nabla_{W_k}f$ according to Eqn.~\eqref{eq:derivation} and we get 
\begin{equation}
\label{eq:based_on}
        \nabla_{W_i}\hat{f} =\sum_{k=i}^{N}\left(\prod_{j=i+1}^{k} W_j\right)^{T} \nabla_{W_k}f\left(\prod_{j=1}^{i-1} W_j \right)^{T}.
\end{equation}

To avoid the complexity of using the standard Lipschitz constant of the smoothness for analysis, we explore and compare the block-wise Lipschitz constant~\cite{BeckT13} for DS-Net and the common network architecture. Specifically, we analyze for each parameter matrix $W_i$ while fixing others. Currently, we have the block-wise Lipschitz constant bound for DS-Net, which is $\| \nabla_{W^1_i}f- \nabla_{W^2_i}f\| \leq L_{i}\left\|W^{1}_{i}-W^{2}_{i}\right\|$. $W^1_i, W^2_i$ are any two possible assignments of $W_i$.

Denote $\lambda_i$ as the largest eigenvalue of the parameter matrix $W_i$, assume we use a 2-norm for the parameter matrix $W_i$, then we get $\lambda_i=\|W_i\|$ where $\lambda_i$ is the largest eigenvalue of $\|W_i\|$. The local smoothness w.r.t. the network parameters of the common network architectures is shown as
\begin{equation}
\footnotesize
\label{eq:convergence_deriv}
\begin{split}
        \left\|\nabla_{W_i^1}\hat{f}-\nabla_{W_i^2}\hat{f}\right\|&=\left\|\sum_{k=i}^{N}\left(\prod_{j=i+1}^{k} W_j\right)^{T}\left(\nabla_{W_k^1}f-\nabla_{W_k^2}f\right)\left(\prod_{j=1}^{i-1} W_j\right)^{T}\right\|\\
                &\leq \sum_{k=i}^{N}\left\|\left(\prod_{j=i+1}^{k} W_j\right)^{T}\left(\nabla_{W_k^1}f-\nabla_{W_k^2}f\right)\left(\prod_{j=1}^{i-1} W_j\right)^{T}\right\| \\
            & \leq \sum_{k=i}^{N}\left(\frac{1}{\lambda_i} \prod_{j=1}^{k} \lambda_j\right) L_k\left\|W^{1}_{k}-W^{2}_{k}\right\| \\
            & \leq\left(\prod_{j=1}^{i-1} \lambda_j\right) L_i \left\|W^{1}_{i}-W^{2}_{i}\right\|. \\
\end{split}
\end{equation}
 The first line of Eqn.~\eqref{eq:convergence_deriv} is obtained based on Eqn.~\eqref{eq:based_on} and the fact $W_j$ is the same when we focus on the investigating the block-wise Lipschitz smoothness of $W_i$. The second line of Eqn.~\eqref{eq:convergence_deriv} is based on triangle inequality of norm. The third line is obtained by the inequality $\|WV\|\leq\|W\|\|V\|$ and the given block-wise smoothness of our DS-Net. The last line is obtained by using the fact that $W_k^1=W_k^2$ if $k\neq i$.
 
 Similarly, for the gradient variance bound of the common network architecture, we start from the gradient variance bound for DS-Net as follows
 \begin{equation}
     \mathbb{E}\left\|\nabla_{W_i}f-\mathbb{E}\nabla_{W_i}f\right\|^{2} \leq \sigma_{i}^{2}.
 \end{equation}
 Given such a bound for DS-Net, the bound for the common network architectures is shown as follows:
 \begin{equation}
 \footnotesize
 \label{eq:variance_deriv}
 \begin{split}
           \mathbb{E}\left\|\nabla_{W_i}\hat{f}-\mathbb{E}\nabla_{W_i}\hat{f}\right\|^{2}&=\mathbb{E}\left\|\sum_{k=i}^{N}\left(\prod_{j=i+1}^{k} W_j\right)^{T}\left(\nabla_{W_k}f-\mathbb{E} \nabla_{W_k}f\right)\left(\prod_{j=1}^{i-1} W_j\right)^{T}\right\|^{2}\\
           &\leq N  \mathbb{E} \sum_{k=i}^{N}\left\|\left(\prod_{j=i+1}^{k} W_j\right)^{T}\left(\nabla_{W_k}f-\mathbb{E} \nabla_{W_k}f\right)\left(\prod_{j=1}^{i-1} W_j\right)^{T}\right\|^{2} \\
           & \leq N\sum_{k=i}^{N}\left(\frac{\sigma_{k}}{\lambda_{i}} \prod_{j=1}^{k} \lambda_{j}\right)^{2}.\\
 \end{split}
 \end{equation}
 The first line of Eqn.~\eqref{eq:variance_deriv} is obtained by using Eqn.~\eqref{eq:based_on}. The second line of Eqn.~\eqref{eq:convergence_deriv} is obtained by Cauchy-Schwarz inequality. The last line is obtained based on the inequality $\|WV\|\leq\|W\|\|V\|$ and the bounded gradient variance of DS-Net.
 \renewcommand{\thesection}{C}
\section{Additional Results on TRADES}
To further illustrate the effectiveness of DS-Net on different adversarial training styles, we test its robustness and standard accuracy on TRADES with both $\beta=1$ and $\beta=6$ in Tab.~\ref{tab:trades}. The detailed experimental setting is described in Section 4.1 of the main paper. Tab.~\ref{tab:trades} shows a similar trend as the DS-Net trained under standard AT and MART as stated in the main paper.   \vspace{-1.5em}
\begin{table*}[ht]
    \centering
    \small
    \tabcolsep 0.04in\renewcommand\arraystretch{0.745}{\small{}}%
    \caption{Evaluations (test accuracy) of deep models on CIFAR-10 and SVHN dataset using TRADES \cite{ZhangYJXGJ19}. $^{1}$ means $\beta=1.0$ and $^{2}$ means $\beta=6.0$. $\dag$ means the results by our implementation. The perturbation bound $\epsilon$ is set to 0.031 for each architecture.}
    \begin{tabular}{c|c|ccccc}
    \hline
      \multicolumn{7}{c}{CIFAR-10}  \\
           \hline
         Defense Architecture&  Param (M)& Natural & FGSM&PGD-20&CW&AA\\
       \hline
         RobNet-large-v2$^{1}$ \cite{GuoYX0L20}$\dag$ &33.42& 87.90$\pm$0.132&57.01$\pm$0.258&49.27$\pm$0.315&49.00$\pm$0.124&46.84$\pm$0.130 \\
        WRN-34-10$^{1}$ \cite{ZagoruykoK16}$\dag$& 46.16&88.07$\pm$0.231 & 56.03$\pm$0.120&49.27$\pm$0.200& 48.98$\pm$0.119&46.62$\pm$0.288 \\
         IE-WRN-34-10$^{1}$ \cite{lmjicml2020}$\dag$ & 48.24&88.31$\pm$0.303 & 54.32$\pm$0.129& 50.22$\pm$0.100& \textbf{50.37}$\pm$0.331& 48.92$\pm$0.138 \\
         \hline
          DS-Net-4-softmax$^{1}$(ours) & 20.78&87.89$\pm$0.176&64.38$\pm$0.218&50.70$\pm$0.322& 46.78$\pm$0.303&48.10$\pm$0.200  \\
                     DS-Net-6-softmax$^{1}$(ours) & 46.35& \textbf{88.44}$\pm$0.301&\textbf{65.00}$\pm$0.120&\textbf{52.50}$\pm$0.286&49.75$\pm$0.174&\textbf{50.00}$\pm$0.166  \\
                     \hline
           Improv.(\%)&-&0.15\%&14.02\%&4.54\%&-&2.21\% \\
                   \hline
                     RobNet-large-v2$^{2}$ \cite{GuoYX0L20}$\dag$ &33.42& 81.95$\pm$0.119&60.31$\pm$0.320&53.21$\pm$0.166&50.09$\pm$0.300&50.13$\pm$0.263 \\
           WRN-34-10$^{2}$ \cite{ZagoruykoK16}$\dag$& 46.16&83.88$\pm$0.110 & 62.28$\pm$0.206&55.49$\pm$0.231& 53.94$\pm$0.158&52.21$\pm$0.145 \\
         IE-WRN-34-10$^{2}$ \cite{lmjicml2020}$\dag$ & 48.24&83.23$\pm$0.134 &61.28$\pm$0.209& 56.03$\pm$0.099&\textbf{61.72}$\pm$0.201& 52.73$\pm$0.273 \\
         \hline
          DS-Net-4-softmax$^{2}$(ours) & 20.78&82.81$\pm$0.375&64.69$\pm$0.154&54.61$\pm$0.272&50.63$\pm$0.219&52.02$\pm$0.116   \\
         DS-Net-6-softmax$^{2}$(ours) & 46.35& \textbf{83.98}$\pm$0.177&\textbf{66.56}$\pm$0.208&\textbf{56.87}$\pm$0.311&54.12$\pm$0.272&\textbf{53.33}$\pm$0.256  \\
         \hline
           Improv.(\%)&-&0.12\%&6.87\%&1.50\%& -&1.14\%\\
          \hline
            \multicolumn{7}{c}{SVHN}  \\
            \hline
             WRN-34-10$^{1}$ \cite{ZhangYJXGJ19}$\dag$& 46.16&94.23$\pm$0.117&72.76$\pm$0.287&52.42$\pm$0.300&48.65$\pm$0.216&48.86$\pm$0.183 \\
              \hline
              DS-Net-4-softmax$^{1}$(ours) & 20.78&94.77$\pm$0.213&72.85$\pm$0.340&\textbf{55.69}$\pm$0.272&48.90$\pm$0.136 &\textbf{51.37}$\pm$0.401 \\
                 DS-Net-6-softmax$^{1}$(ours) & 46.35&\textbf{95.73}$\pm$0.197&\textbf{76.61}$\pm$0.362&54.92$\pm$0.351& \textbf{49.12}$\pm$0.272&51.26$\pm$0.228  \\
                 \hline
           Improv.(\%)&-&1.59\%&5.29\%& 6.24\%&0.97\%&5.14\%\\
                 \hline
                    WRN-34-10$^{2}$ \cite{ZhangYJXGJ19}$\dag$& 46.16&91.92$\pm$0.223&73.65$\pm$0.128&57.46$\pm$0.125&50.34$\pm$0.231&54.11$\pm$0.231 \\
                    \hline
              DS-Net-4-softmax$^{2}$(ours) & 20.78&91.74$\pm$0.370&\textbf{73.83}$\pm$0.414&59.84$\pm$0.351&53.92$\pm$0.184&56.54$\pm$0.306  \\
              DS-Net-6-softmax$^{2}$(ours) & 46.35&\textbf{92.54}$\pm$0.217&73.04$\pm$0.361&\textbf{60.54}$\pm$0.212&\textbf{54.58}$\pm$0.153&\textbf{56.75}$\pm$0.065\\
              \hline
           Improv.(\%)&-&0.67\%&0.24\%&5.36\%&8.42\%&4.88\%\\
           \hline
    \end{tabular}
    \label{tab:trades}
\end{table*}
 \renewcommand{\thesection}{D}
\section{Effect of Weight Decay}
To demonstrate the effect of weight decay in adversarial training, we change the weight cay to 3e-4, 4e-4, 6e-4 and 7e-4 and report the average performance in terms of robustness and generalization ability for DS-Net. We conduct experiments using standard AT on CIFAR-10 with factor $k=4$, which are shown in Tab.~\ref{tab:weight_decay}. The results demonstrate the importance of weight decay in adversarial training (align well with the empirical findings in~\citet{pang2021bag}), which should be carefully selected.
\begin{table}[!h]
    \centering
    \small
     \tabcolsep 0.04in\renewcommand\arraystretch{0.745}{\small{}}%
         \caption{Model robustness and generalization ability with different weight decay. The perturbation bound for evaluation is set to 0.031.}
    \begin{tabular}{c|ccccc}
    \hline
        Weight decay & 3e-4&4e-4&5e-4&6e-4&7e-4 \\
         \hline
      PGD-20 Acc. (\%)&48.28 &49.69  &\textbf{ 54.14}& 52.67&49.69\\
      Standard Acc. (\%)& 82.50&85.00 &\textbf{ 85.39}& 84.06&83.75 \\
         \hline
    \end{tabular}
    \label{tab:weight_decay}
\end{table}
 \renewcommand{\thesection}{E}
\section{Comparison with using different optimizers for DS-Net}
SGD is commonly used in AT literature. We tried other optimizers such as Adam, RMSprop, Adadelta and Adagrad. The PGD-20 accuracy is listed in Tab.~\ref{tab:optimizer} for DS-Net-4-softmax on CIFAR-10 (vs. 54.14\% by SGD).
\begin{table}[!h]
    \centering
    \small
    \caption{ PGD-20 accuracy of DS-Net trained with different optimizers for block parameters.}
    \begin{tabular}{c|cccc}
    \hline
       Optimizer  & Adam&Adadelta&Adagrad&RMSprop \\
       \hline
        PGD-20 Acc. &40.76\%& 41.12\%& 39.28\%&37.46\% \\
         \hline
    \end{tabular}

    \label{tab:optimizer}
\end{table}





\end{document}